\pdfoutput=1

\documentclass[11pt]{article}

\usepackage[]{acl}

\usepackage{times}
\usepackage{latexsym}

\usepackage[T1]{fontenc}

\usepackage[utf8]{inputenc}

\usepackage{microtype}

\usepackage{inconsolata}

\usepackage{CJKutf8}

\usepackage{hyperref}
\usepackage{pdfpages}
\usepackage{fancyvrb}
\usepackage{tcolorbox}
\usepackage{amsmath}
\usepackage{booktabs,arydshln}
\usepackage{multirow}
\usepackage{graphicx}
\usepackage{tabularx}
\usepackage{color, colortbl}
\definecolor{bblue}{rgb}{0.0, 0.5, 1.0}
\definecolor{pcolor}{rgb}{0,0.5,0.5}
\definecolor{gcolor}{rgb}{0.0, 0.5, 1.0}
\definecolor{myblue}{rgb}{50,120,200}
\definecolor{mydarkblue}{rgb}{0,0,139}

\newcommand{\TODO}[1]{\textcolor{blue}{}}
\newcommand{\prompt}[1]{{\textit{``#1''}}}

\definecolor{purp}{HTML}{791f87}

\makeatletter
\renewcommand{\sectionautorefname}{Section \@gobble}
\renewcommand{\subsectionautorefname}{Section \@gobble}
\renewcommand{\subsubsectionautorefname}{\S\@gobble}
\renewcommand{\appendixautorefname}{Appendix \@gobble}
\makeatother

\newcommand{\lm}[1]{\textsc{#1}}

\title{Does Liking Yellow Imply Driving a School Bus? \\ Semantic Leakage in Language Models}

\author{Hila Gonen$^{1}$ \quad Terra Blevins$^{1}$ \quad Alisa Liu$^{1}$ \quad Luke Zettlemoyer$^{1}$ \quad Noah A.~Smith$^{1,2}$ \\
        $^{1}$Paul G. Allen School of Computer Science \& Engineering, University of Washington \\ 
        $^{2}$Allen Institute for Artificial Intelligence \\
        {\tt hilagnn@gmail.com}\\
        {\tt \{blvns,alisaliu,lsz,nasmith\}@cs.washington.edu}\\}

\usepackage{fancyhdr}
\fancypagestyle{firststyle}
{
    \fancyhf{}
    \fancyhead{}
    \lhead{Published in NAACL 2025}
}

\begin{document}
\maketitle
\thispagestyle{firststyle} 
\begin{abstract}

Despite their wide adoption, the biases and unintended behaviors of language models remain poorly understood. In this paper, we identify and characterize a phenomenon never discussed before, which we call \textit{semantic leakage}, where models leak irrelevant information from the prompt into the generation in unexpected ways. We propose an evaluation setting to detect semantic leakage both by humans and automatically, curate a diverse test suite for diagnosing this behavior, and measure significant semantic leakage in 13 flagship models. We also show that models exhibit semantic leakage in languages besides English and across different settings and generation scenarios. This discovery highlights yet another type of bias in language models that affects their generation patterns and behaviour.

\end{abstract}

\section{Introduction}

As language models (LMs) become more prevalent~\cite{llama2,gemini,gpt4, bloom}, we are steadily learning more about their peculiarities and the unique and often unexpected properties of their behavior. Phenomena ranging from hallucinations~\cite{hallucination-survey} to sycophancy~\cite{sycophancy} and many types of biases~\cite{bias-survey} have been revealed in these models' outputs. Each such discovery leads to a cycle of in-depth study and development of new methods to mitigate these behaviors as much as possible.  

\begin{figure}[ht]
    \centering
    \resizebox{0.9\linewidth}{!}{
    \tcbset{colback=blue!8!white, colframe=bblue!75!black, 
    fonttitle=\bfseries, title=Examples of semantic leakage, width=\linewidth+0.1cm, halign title=center }
    
    \begin{tcolorbox}
    
    Prompt: \textcolor{pcolor} {\textit{He likes \underline{yellow}. He works as a}} \\
    \lm{Gpt4o}: \textcolor{gcolor} {\textit{school bus driver}} \\ 

    Prompt: \textcolor{pcolor} {\textit{He likes \underline{ants}. His favorite food is}} \\
    \lm{Gpt4o}: \textcolor{gcolor} {\textit{ant-covered chocolate, a unique delicacy that combines the crunch of ants with the sweetness of chocolate}} \\ 
    
    Prompt: \textcolor{pcolor} {\textit{It was her turn to speak and she \underline{got cold feet}. A day later she stayed at home because she got hurt in her}} \\
    \lm{Gpt4o}: \textcolor{gcolor} {\textit{ankle}} \\ 
    
    Prompt: \textcolor{pcolor} {\textit{He is a \underline{doctor}. His favorite song is}} \\
    \lm{Gpt4o}: \textcolor{gcolor} {\textit{'Stayin' Alive' by the Bee Gees}} \\

    \end{tcolorbox}  
    }
    \caption{Examples of semantic leakage in \lm{Gpt4o}. The leaking concept is underlined.}  
    \label{fig:examples}
\end{figure}

We identify a phenomenon in language models never discussed before, which we term \textbf{semantic leakage} --- these models can generate text with strong semantic relationships to unrelated words in the prompts. For example, when given the prompt \prompt{He likes \textbf{yellow}. He works as a}, \lm{Gpt4o}\footnote{\url{https://platform.openai.com/docs/models/gpt-4o}} generates the output \prompt{\textbf{school bus driver}} (\autoref{fig:examples}). Here we say that the word \textit{yellow} has leaked into the generation in a way that unintentionally influences the generated occupation. Figure~\ref{fig:examples} shows three additional examples of prompt-generation pairs (using \lm{Gpt4o}). In each example, the leakage from the semantic meaning of the underlined word in the prompt is apparent in the generation.

We define semantic leakage in a generation as an undue influence of semantic features from words in the prompt on the generation, ``undue'' in the sense that the semantic relatedness between the prompt and the generation is stronger than would be expected in natural distributions. Often semantic leaks read as forced, overwrought, even nonsensical generations, like those found in children's stories.

In this paper, we introduce an evaluation metric for measuring semantic leakage. We examine semantic leakage with 109 examples of different semantic categories (animals, food, music, etc.) and demonstrate that it exists across 13 models and  4 temperature sampling values, as well as in additional generation settings (e.g., open-ended generation and multilingual settings). Our analysis shows that finetuned/instruction-tuned models tend to leak \emph{more}, and that semantic leakage also happens  across languages.

Semantic leakage is closely related to different types of biases models exhibit, ranging from gender, racial and cultural biases~\cite{bolukbasi,caliskan,lipstick,stereoset} to cognitive and psychological biases~\cite{cog-bias-codex,cog-bias-llms,reasoning-bias-nature}, in which associations between different concepts are learned by the model during training and exposed as bias during generation \cite{mt-bias,self-debias}. 
As an example, consider the prompt \prompt{\textbf{She} works at the hospital as a}, and the prompt \prompt{\textbf{He} works at the hospital as a}. Given the generations \prompt{nurse} and \prompt{doctor}, respectively, which is a typical biased behavior, we can think of the word ``she'' as the concept that  leaks the property of the female gender into the generation \prompt{nurse}, a stereotypically female occupation.
While still not fully understood, we suspect that much documented and discussed gender bias and other types of previously documented biases \cite{bias-survey} are instances of associations that get learned and influence in a broader way, which is partially reflected as semantic leakage. 
Specifically, here we are looking at larger semantic classes (i.e., compared to gender or race), and we seek to quantify and study learned associations and their effect on model generation settings.

Given the nature of models in learning associations during training, whether semantic leakage is surprising or not is a point of contention. In addition, the implications of this behaviour depend on the user and application context, and are not necessarily either good or bad. At the same time, we strongly believe that it is an interesting and important behavior to investigate as it may be a broad enough umbrella to encompass many other associations that are studied as more specific cases. Studying the broader family may be easier and more beneficial, as the more general class may include cases more resistant to mitigation strategies and more prevalent in model outputs.

Finally, in addition to typical and previously discussed biases of different types that fall under this broad class of semantic leakage, we point to additional potential ramifications of this phenomenon. First, semantic leakage might hinder the performance of models via the overshadowing mechanism, introduced in \citet{overshadow}, in which strong associations in a question override more important and relevant parts of the question to a point where the latter are ignored, resulting in a wrong answer. Semantic leakage might also be used in adversarial settings, similar to attacks presented in \citet{universal-adversarial}, where specific words that are added to the prompt negatively influence the output of the model. Other potential implications include hindering performance in creative writing \cite{creative}, where diversity and originality are especially important, but semantic leakage might significantly hurt the output quality in unexpected ways.

Our contributions in this paper can be summarized as follows: (1) we identify and define the phenomenon of \textit{semantic leakage} in language model generation (\autoref{sec:semantic}); (2) we build a test suite for detecting semantic leakage in language models (\autoref{sec:semantic}); (3) we evaluate 13 models with varying sizes using this test suite, uncovering consistent cross-model trends, and validate this automatic evaluation with human judgments (\autoref{sec:auto-results} and~\ref{sec:human-eval}); (4) we show that models also exhibit semantic leakage in languages beyond English (Chinese and Hebrew) as well as in crosslingual settings (\autoref{sec:multi}) and in more open-ended generation (\autoref{sec:open}). By characterizing semantic leakage, we demonstrate yet another property language models exhibit in text generation, and highlight how choices in prompt construction can inadvertently affect model output.

\section{Semantic Leakage}
\label{sec:semantic}

\subsection{Overview and Definitions}

When producing text, language models can draw on semantic associations with words from the input, or prompt, that are not required or expected, and sometimes even violate rules of logic or common sense. For example, given the prompt \prompt{He likes \textbf{koalas}. His favorite food is} \lm{Gpt4o} generates the output \prompt{\textbf{eucalyptus leaves}}. Here, we say that the semantic association with ``koalas'' and the foods they eat ``leaks'' into the generation, despite the fact that a person's favorite food and their opinion on koalas are unrelated in the real world. We call this phenomenon \textbf{semantic leakage}.  

A related phenomenon of conceptual leakage has been reported in image generation \cite{dalle-leakage}. There, they find that visual properties of one object leak into other objects in the image (for example, the prompt ``\textit{a zebra and a street}'' generates an image of a zebra next to a zebra crossing), which resembles examples we show (e.g., yellow leaking into the occupation through ``school bus driver'', see first example in~\autoref{fig:examples}). While this behaviour might be seen as subjective, the cases we consider in this paper are, we believe, beyond debate.

Semantic leakage in text generation can also manifest in more subtle ways: for the prompt \prompt{He likes \textbf{green}. He works as a}, \lm{Gpt4o} generates \prompt{\textbf{landscape architect}} as a response. In other cases, the model may leak semantics that are not even used in the prompt: For example, when prompted with an idiom, a model can leak the literal semantic meaning of that phrase (that is not actually being used): for instance, when prompted with \prompt{She gave him the \textbf{green light} for the new project. A day later he sent an invitation to everyone by mail, with an envelope colored}, \lm{Gpt3.5}  generates the response \prompt{bright green to match the theme of the project.} This is similar to another observation made by \citet{dalle-leakage} where the authors demonstrate that sense-ambiguous words are hard for the model to isolate, and the generated images often exhibit the unintended sense together with the intended one.

\subsection{Operationalizing the Measurement of Semantic Leakage}
\label{sec:operation}

We define the different elements of semantic leakage as follows. A \textit{prompt} is input text that primes the model to output a corresponding \textit{generation}. We consider two types of prompts: \textit{control} prompts, which do not include any spurious semantic signal (\prompt{His favorite food is}), and \textit{test} prompts (\prompt{He likes \textbf{koalas}. His favorite food is}), which mirror the control prompt but add a semantically unrelated \textit{concept} (``koalas'') to the input, leading to a different, \textit{test generation}. While it is known that changing the surface form of the prompt often alters model output \cite{demystifying,prompt_format}, these new \textit{test} generations are frequently much more semantically similar to the \textit{concept} than the \textit{control} generations (\autoref{sec:auto-results}).

We evaluate the prevalence of semantic leakage in a given model by comparing the similarity of the generations produced by the control and test prompts to the concept under consideration. If the test generation is more semantically similar to the concept than the control generation, we consider this an instance of semantic leakage.

To quantify the prevalence of semantic leakage, we design an evaluation setting that is motivated by the definition of semantic leakage, as having stronger connections of the concept to the test generation. The goal is to compare the similarity of the generations produced by the control and test prompts to the concept, while making sure no other factors are taken into consideration. We use the following similarities to then derive the evaluation metric detailed below:

\begin{align*}
    \text{sim}_{\text{control}} & = similarity(\text{concept}, \text{control}) \\
    \text{sim}_{\text{test}}  & = similarity(\text{concept}, \text{test})
\end{align*}

\paragraph{Evaluation Metric}
\label{sec:eval-metrics}

From the above formulation, we derive the ``Semantic Leakage Rate'' metric (\textbf{Leak-Rate}), the percentage of instances in which the concept is semantically closer to the test generation than the control generation. We score \textit{Leak-Rate} by averaging the following function across all instances, and converting to the range of 0--100\%:

\begin{align}
\label{eq:leak-rate}
\resizebox{\columnwidth}{!}{
$
\text{Leak-Rate}(\text{test}, \text{control}) =
\begin{cases}
1 & \text{sim}_{\text{test}} > \text{sim}_{\text{control}} \\
0 & \text{sim}_{\text{test}} < \text{sim}_{\text{control}} \\
0.5 &  \text{sim}_{\text{test}} = \text{sim}_{\text{control}} 
\end{cases}
$
}
\end{align}
This evaluation setting can be implemented as automatic evaluation (\autoref{sec:auto-results}) or as human evaluation (\autoref{sec:human-eval}), with no required adjustments. For automatic evaluation, we embed the texts (\textit{concept, control generation, test generation}) with an embedding method (see~\autoref{sec:embed}) and compute similarity measures; for human evaluation, we ask annotators to judge which of the two pairs is more semantically similar.

When the model does not exhibit semantic leakage at all, we expect a Leak-Rate of 50\%, i.e., an even split between test vs.~control having higher similarity, for those that show a difference. We expect Leak-Rate higher than 50\% when the model exhibits semantic leakage. The Leak-Rate scores are comparable across models for a given embedding method and test set of prompt pairs.

\subsection{Building a Test Suite}

We build a test set of prompts containing categories and concepts with clear semantic associations. We also focus on categories and prompt settings that are likely to allow uncontroversial evaluation (by priming for short model outputs), though later we also consider open-ended generation (\autoref{sec:open}). In addition, a subset of the test suite considers idioms, which have both literal and figurative interpretations, as concepts in the test prompts.

We manually create 109 prompts with concepts from categories such as colors, food, animals, songs, occupations and more. A few examples are listed in \autoref{tab:examples}.\footnote{The full list can be found at \url{https://gonenhila.github.io/files/semantic_leakage_prompts.txt}} Each prompt in our test suite is matched with a control prompt as explained above. When designing the control prompts, we prioritized keeping them as basic and simple as possible to avoid confounding factors. In some cases, more elaborate prompts required us to include richer control prompts that are more similar to the test prompts in their length and structure.\footnote{An example is the prompt \prompt{She went on a field trip to the zoo. Her sister got a gift from a friend: a shirt with a print of} for which we design the control prompt: \prompt{She went on a field trip. Her sister got a gift from a friend: a shirt with a print of}.} Both types of control prompts are included in our evaluation.

\begin{table*}[ht]
\centering\small
\resizebox{0.95\linewidth}{!}{
\begin{tabular}{p{1.5cm} p{14cm}}
\toprule
\textbf{Model} & \textbf{Prompt and generation} \\\midrule
\lm{Llama2} & \textit{\textcolor{pcolor} {John likes \underline{dolphins}. John's father is working as a}} \textcolor{gcolor} {marine biologist} \\ \cline{2-2}
\lm{7b-chat} & \textit{\textcolor{pcolor} {John likes \underline{tigers}. John's father is working as a} } \textcolor{gcolor}{tiger in a zoo} \\ \midrule

\lm{Llama2} & \textit{\textcolor{pcolor} {He likes \underline{pandas}. His favorite food is}} \textcolor{gcolor} { bamboo} \\ \cline{2-2}
\lm{13b-chat} & \textit{\textcolor{pcolor} {He likes \underline{mice}. His favorite food is}} \textcolor{gcolor} { cheese } \\ \midrule

\lm{Llama2} & \textit{\textcolor{pcolor} {His name is \underline{Cedar}. His friend lives in}  } \textcolor{gcolor} {a treehouse} \\ \cline{2-2}
\lm{70b-chat} & \textit{\textcolor{pcolor} {His name is \underline{Ocean}. His friend lives in}  } \textcolor{gcolor} {a houseboat} \\ \midrule

\lm{Llama3} & \textit{\textcolor{pcolor} {He watched a \underline{polka} concert. For dinner he had}} \textcolor{gcolor} {a plate of pierogies and a side of sauerkraut} \\ \cline{2-2} 
\lm{8b-ins} & \textit{\textcolor{pcolor} {He watched a \underline{flamenco} concert. For dinner he had}} \textcolor{gcolor} {a delicious paella} \\ \midrule

\lm{Llama3} & \textit{\textcolor{pcolor} {\underline{Mercedes} has a friend. The friend's hobby is} } \textcolor{gcolor} { collecting Mercedes cars because he loves the Mercedes brand} \\ \cline{2-2} 
\lm{70b-ins} & \textit{\textcolor{pcolor} {\underline{Forest} has a friend. The friend's hobby is } } \textcolor{gcolor} { gardening, to her home in a way which makes it a forest} \\ \midrule

\multirow{2}{*}{\lm{Gpt3.5}}  & \textit{\textcolor{pcolor} {He is a \underline{firefighter}. His favorite song is }} \textcolor{gcolor} {``Ring of Fire'' by Johnny Cash} \\ \cline{2-2}
 & \textit{\textcolor{pcolor} {He is a \underline{postman}. His favorite song is }} 
\textcolor{gcolor} {``Signed, Sealed, Delivered'' by Stevie Wonder} \\ \midrule

\multirow{4}{*}{\lm{Gpt4}}  & \textit{\textcolor{pcolor} {The woman was \underline{as cold as ice}. Afterwards, I went to the library and read }} 
\textcolor{gcolor} {about the fascinating science of cryogenics, which explores the process of freezing and preserving organic matter at extremely low temperatures} \\ \cline{2-2}
 & \textit{\textcolor{pcolor} {It was her turn to speak and she \underline{got cold feet}. A day later she stayed at home because she got hurt in her }} 
\textcolor{gcolor} {ankle during a morning jog} \\\midrule

\multirow{2}{*}{\lm{Gpt4o}}  & \textit{\textcolor{pcolor} {She is \underline{a music lover}. Her nephew was accused of } }
\textcolor{gcolor} {stealing her prized vinyl collection} \\ \cline{2-2}
 & \textit{\textcolor{pcolor} {She is \underline{a cinema lover}. Her nephew was accused of }} 
\textcolor{gcolor} {pirating movies} \\

\bottomrule
\end{tabular}
}
\caption{Examples of instances that lead to semantic leakage, taken from different models. The prompt is in green (\textit{italic}) and the generation is in blue.}
\label{tab:examples}
\end{table*}

\section{Experimental Setup}

We present the experimental setup for automatically evaluating semantic leakage in a wide range of models and model sizes.

\subsection{Experimental Details}
\label{sec:exp}

We evaluate semantic leakage in multiple language models from two families: \lm{GPT}\footnote{\url{https://platform.openai.com/docs/models}} and \lm{Llama} models \cite{llama2}, as detailed below. For all models, we explore several temperature values (0, 0.5, 1, 1.5), and run each prompt 10 times to get variation in the generations, when possible.

Before evaluation, for cases where the prompt is repeated in the model generation, we remove the repeated prompt. We also truncate the generations after the first period since the main piece of information is generated before it, and because \lm{Llama} models tend to generate unrelated sentences or phrases that might interfere with the evaluation.

\subsection{Models}

We experiment with 13 models of two different  state-of-the-art model families to explore semantic leakage in a diverse inventory of models.

\paragraph{GPT models}

We use OpenAI's API and send requests to \lm{GPT} models by calling \lm{Gpt3.5}, \lm{Gpt4}~\cite{gpt4}, and \lm{Gpt4o}.\footnote{\texttt{gpt-3.5-turbo-0125}, \texttt{gpt-4-turbo-2024-04-09} and \texttt{gpt-4o-2024-05-13}, respectively.}
With \lm{GPT} models, for sentence completion prompts we prepend ``Complete the sentence:'', as we find the model performs the task better this way.

\paragraph{\lm{Llama} models}

We run all \lm{Llama} variations using Huggingface \cite{huggingface}.\footnote{\lm{Llama2}: 7B, 7B-chat, 13B, 13B-chat, 70B, 70B-chat. \lm{Llama3}: 8B, 8B-Instruct, 70B, 70B-Instruct.}
We cap the generation in \lm{Llama} models at 100 tokens (300 tokens for open generation, \autoref{sec:open}).

\subsection{Embedding Methods}
\label{sec:embed}

For automatic evaluation, we aim to use basic embedding methods that are able to detect and reflect semantic similarities, and are ideally detached from the models we evaluate to avoid confounding factors. We consider the following embedding methods. \textsc{BERT-score} directly provides a similarity score, for the others we apply cosine-similarity.

\paragraph{\textsc{BERT-score} (BS)}

\textsc{BERT-score} \cite{bertscore} is an automatic evaluation metric for text generation, that computes a similarity score for each token in the candidate sentence with each token in the reference sentence, where token similarity is computed using contextual embeddings. We use the \texttt{distilbert-base-uncased} model.\footnote{https://huggingface.co/spaces/evaluate-metric/bertscore} 

For the multilingual experiments, where we expect generations in non-English languages (\autoref{sec:multi}), we use the respective models: \texttt{bert-base-chinese} for Chinese, and \texttt{bert-base-multilingual-cased} for Hebrew. Crosslingual settings still use the English model as the generations there are mainly in English.

\paragraph{\textsc{SentenceBERT embeddings} (SB)}

\textsc{SentenceBERT} \cite{sentence-bert} is a modification of BERT \cite{bert} that uses Siamese and triplet network structures to derive semantically meaningful sentence embeddings that can be compared using cosine-similarity. We use the huggingface implementation.\footnote{https://huggingface.co/efederici/sentence-bert-base}

\paragraph{\textsc{OpenAI embeddings} (OAI)}

We use \texttt{text-embedding-3-large},\footnote{\url{https://platform.openai.com/docs/guides/embeddings/embedding-models}} OpenAI's best performing embeddings. There is no public documentation of the model training.

In addition to these metrics, we validate our experiments with a manual evaluation of semantic leakage on a subset of the models (Section \ref{sec:human-eval}). This also serves as a validation of our automatic metrics.

\section{Results}
\label{sec:auto-results}

\paragraph{Significant semantic leakage across various use cases and models.}
\autoref{tab:all} depicts the average leakage for each model, across multiple samplings and temperature values, as detailed in \autoref{sec:exp}. We see that semantic leakage is exhibited by all model variations, and is detected by all embedding models we use. Leak-Rate values are all well above the 50\% random mark and statistically significant\footnote{Using a $t$-test for the lists of the Leak-Rate values to test that their mean is significantly greater than 50\%.} with $p<10^{-100}$, validating the semantic relatedness of the prompt and the test generation.

\begin{table}[ht]

\centering

\resizebox{0.65\linewidth}{!}{
\begin{tabular}{l|r|r|r}

& \multicolumn{3}{c}{\textbf{Leak-Rate}} \\
\textbf{Model} & BS & SB & OAI\\ \hline \hline

GPT3.5 & 74.3 & 68.6 & \textbf{85.5} \\
GPT4 & 70.8 & 61.2 & 84.4 \\
GPT4o & \textbf{76.9} & \textbf{70.4} & 85.0 \\ \midrule
2-7b & 66.8 & 64.9 & 72.8 \\
2-7b-chat & 72.6 & \textbf{71.7} & 77.8 \\
2-13b & 70.4 & 65.1 & 73.6 \\
2-13b-chat & 71.5 & 65.2 & \textbf{78.4} \\
2-70b & 71.9 & 63.2 & 75.9 \\
2-70b-chat & \textbf{75.4} & 66.8 & 78.2 \\ \midrule
3-8b & 69.6 & 65.9 & 75.5 \\
3-8b-ins & \textbf{78.1} & 68.8 & \textbf{81.5} \\
3-70b & 71.6 & 68.1 & 75.2 \\
3-70b-ins & 76.3 & \textbf{71.2} & 77.3 \\

\bottomrule
\end{tabular}
}
\caption{Semantic Leak-Rate averaged across 10 samples for each of 4 temperature values. No semantic leakage would correspond to a Leak-Rate of 50\% (random guessing), with higher values indicating more leakage. The bottom sections present the \lm{Llama}2 and \lm{Llama}3 models, respectively. The model showing the most leakage in each <model family, metric> setting is \textbf{bolded}.}
\label{tab:all}
\end{table}

\autoref{tab:examples} lists a few examples of semantic leakage from the different models, showcasing leakage in diverse use cases and styles and with respect to a variety of leaking concepts. In many cases we explore, the generations do not make sense in the context, or are very limited and focused on the leaking concept from the prompt.

\paragraph{Leakage is more pronounced in certain model variations.}
The results in~\autoref{tab:all} show that certain model variations tend to exhibit more semantic leakage than their counterparts. We now analyze the differences within the 2 model families.

\paragraph{GPT models}
\autoref{fig:app_gpt4o} in the Appendix shows the leakage estimation in the three different \lm{GPT} models across different temperature values. \lm{Gpt4o} consistently leaks more than \lm{Gpt4} and \lm{Gpt-3.5}.\footnote{This is not true for the OpenAI embedding model, which may be due to confounding model training factors; however, how the embeddings are constructed is not publicly available.} 

\begin{figure}[ht]
\centering
    \includegraphics[width=\columnwidth,trim=2em 1em 2em 1em]{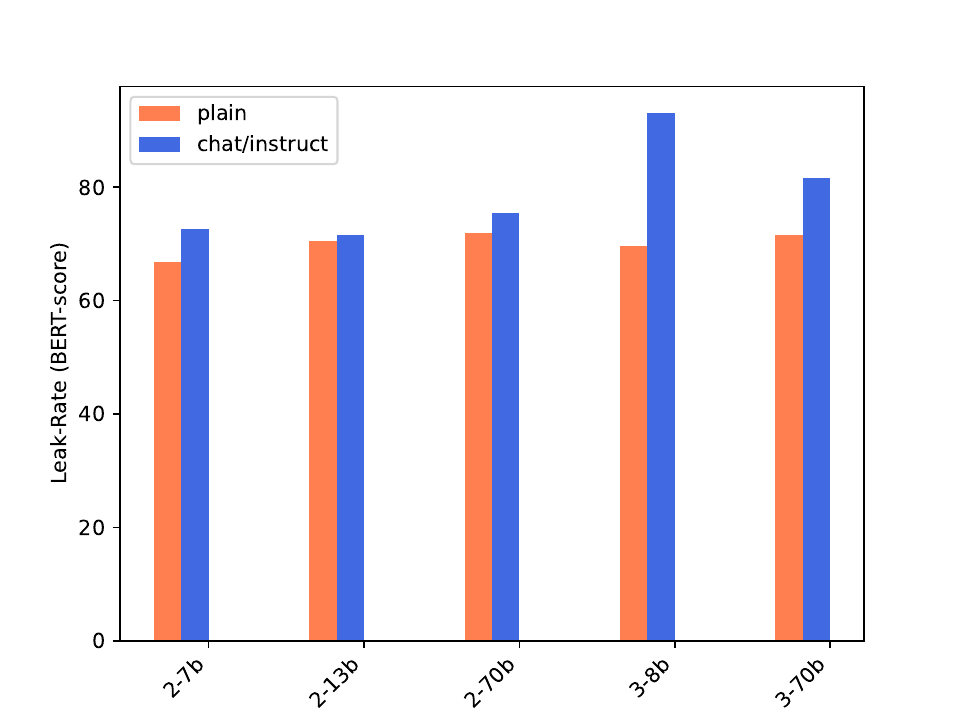}
    \caption{Semantic leakage in \lm{Llama} models, averaged across temperature values (measured with Leak-Rate using BERT-score). }
    \label{fig:llama_chat}
\end{figure}

\paragraph{\lm{Llama} models}
For \lm{Llama} models we consistently see that the instruction-tuned models (\textsc{chat} version in \lm{Llama}2 and \textsc{instruct} version in \lm{Llama}3) leak more than their pretrained-only counterparts. A detailed comparison is presented in \autoref{fig:llama_chat} where we plot the average leakage of each model (averaged across temperature values), as measured with Leak-Rate with BERT-score embeddings. All the differences are statistically significant\footnote{$t$-tests on the Leak-Rate values show that the mean in finetuned models is significantly higher than vanilla versions.} with $p < 0.002$ except for Llama-2-13b. We see similar trends with all other metrics as well.

\paragraph{Leakage across different temperatures.}

We now inspect the way sampling temperature affects semantic leakage.
For the \lm{GPT} models, we see no clear trends (\autoref{fig:app_gpt4o} in the Appendix).\footnote{It is unclear whether temperature behaves as expected in the \lm{GPT} API. We note that a temperature setting of zero (which should mean greedy, deterministic decoding) will give different outputs on repeated calls to the API.}

\begin{figure}[ht]
\centering
    \includegraphics[width=\columnwidth,trim=2em 1em 2em 1em]{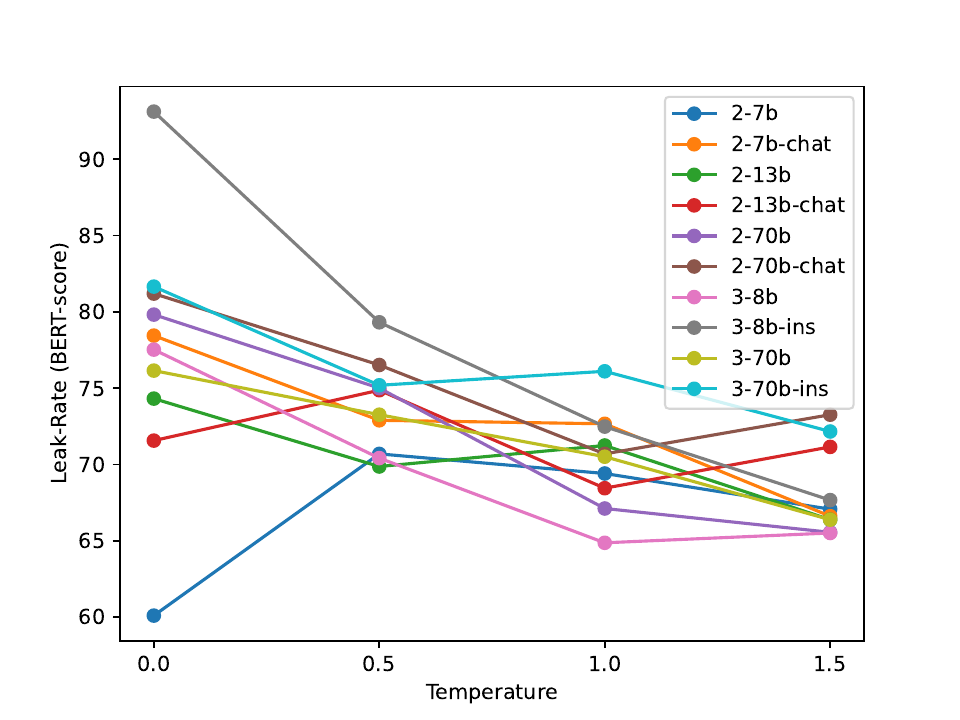}
    \caption{Semantic leakage in \lm{Llama} at different temperatures (measured with Leak-Rate using BERT-score). }
    \label{fig:llama_temp}
\end{figure}

For \lm{Llama} models, we see that greedy sampling ($t=0$) leads to the highest semantic leakage measures (see \autoref{fig:llama_temp}). Generally, lower temperature values lead to more leakage---this is consistent for most models and across all metrics.

\section{Human Evaluation}
\label{sec:human-eval}

We now perform a manual evaluation of semantic leakage, which will also validate our automatic metrics and experiments. 

\paragraph{Human Evaluation Setup}

Human evaluation is based on the same setting as automatic evaluation, described in \autoref{sec:operation}.

We recruit two native English speakers who are not involved with the project. These annotators are not provided with the objective of this experiment, but instead are given the following annotation guidelines: \textit{Consider the word or phrase X. Which of the following texts (A or B) is more \textbf{semantically related} to X? (A/B/Neither)}. We map the annotator's choice of (A/B/Neither) to (test, control, neither) and then score Leak-Rate (Eq.~\ref{eq:leak-rate}).

Using these guidelines, the annotators are asked to label 109 test-control generation pairs from each model (ordered randomly). For the human evaluation, we consider the largest model from each model family at the temperature $t$ found to leak the most by automatic metrics: \lm{Gpt4o} ($t=1$) and \lm{Llama}3-70B Instruct ($t=0$).

\paragraph{Analysis}

\begin{figure}[ht]
\centering

    \includegraphics[scale=0.49]{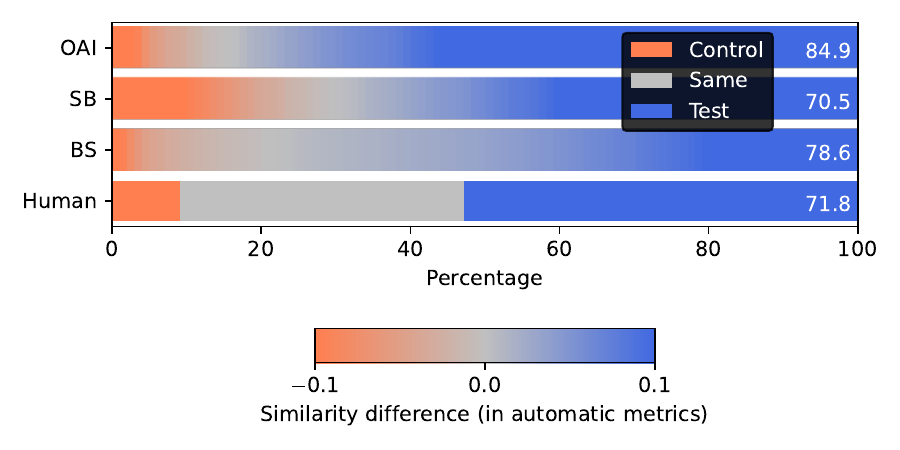}

    \caption{Human detection of semantic leakage compared to automatic methods. Leak-Rate is reported on the right for each method.}
    \label{fig:human_eval_en_gradient}
\end{figure}

\autoref{fig:human_eval_en_gradient} compares the semantic leakage detected by the human evaluation for \lm{Gpt4o} against the automatic metric. 
The values in the human evaluation row are percentages for each category: test is more similar to the concept, control is more similar to the concept, or neither. 

We expect humans to have higher tolerance for similar scores, i.e., more cases falling under $\text{sim}_{\text{test}} = \text{sim}_{\text{control}}$ in the human evaluation than in the automatic evaluation, where it occurs almost solely when the test and control generations are the same (see \autoref{eq:leak-rate}). To visually account for this difference we plot the automatic metric results by using colored gradient to depict the difference in similarity values $\text{sim}_{\text{test}} - \text{sim}_{\text{control}}$, with positive values (shades of blue on the right) implying semantic leakage. The results account for all ten generations sampled from each model. 

We find similar trends on \lm{Llama}3-70B Instruct with an average human-annotated Leak-Rate of 66.7 and automatic evaluation of Leak-Rate that range from 71.2 to 77.3 across the different embedding types.

We also calculate Kendall's $\tau$ on the human annotations. We find high interannotator agreement between the human annotators ($\tau = 0.68$), indicating that humans generally agree on the cases that constitute semantic leakage. We also compute Kendall's $\tau$ on the human evaluation vs. similarity differences calculated using BERTScore embeddings to evaluate how well these methods correlate,\footnote{We introduce a slack variable of $\epsilon$ = 0.03 to account for the different levels of similarity tolerance between humans and embedding-based similarity.} and get a moderate correlation of $\tau = 0.39$ when averaged between the two annotators.\footnote{While we use all model generations per example when obtaining label distributions and computing Leak-Rate, to calculate $\tau$ we use the generations shown to the human annotators, as this requires example-level alignments.} 

\section{Multilingual and Crosslingual Semantic Leakage}
\label{sec:multi}

In the previous sections we established that semantic leakage is exhibited in various scenarios in English, across different models and model sizes. We now inspect semantic leakage in languages outside of English, as well as in crosslingual settings, by focusing on Hebrew and Chinese, Hebrew-English and Chinese-English. As in the previous section, we consider the largest model from each model family at the temperature $t$ found to leak the most: \lm{Gpt4o} ($t=1$) and \lm{Llama}3-70B Instruct ($t=0$). Since we found that \lm{Llama} is struggling with Hebrew, especially with Hebrew generation, we exclude these results.

\paragraph{Multilingual Semantic Leakage}

We translate all the prompts into Chinese and Hebrew (see examples in~\autoref{tab:app_examples_multi} in the Appendix, rows 1, 3) with the help of native speakers (including the authors). The translation is mostly straightforward with the exception of idioms and names. For English idioms that do not exist in the target language, we write a new example using an idiom in that language (together with a matching context and a control prompt). For names, in Chinese we choose a plausible Chinese name with that meaning (e.g., \begin{CJK*}{UTF8}{gbsn}
小麦
\end{CJK*}, meaning \textit{little wheat}, for Rye). In Hebrew, we make an attempt to translate to an existing frequent name in Hebrew, or to substitute it with a name with a related meaning (e.g., Lake instead of River). If that is not an option (e.g., Rye), we come up with a different name that has a clear semantic field.

\paragraph{Crosslingual Semantic Leakage}

We create mixed prompts that begin with Chinese/Hebrew and end with English (see examples in~\autoref{tab:app_examples_multi} in the Appendix, rows 2, 4). We use the translations to Chinese and Hebrew, and mix with English: for the first part of the prompt we use the translation to Chinese/Hebrew, and for the rest of it we use the original part in English. For evaluation purposes, we use the English concepts, since we expect the generation to be in English. For names or idioms that are used in Chinese/Hebrew parts of the prompt, we use their literal translations into English.

\paragraph{Results}

\begin{figure}[ht]
\centering

    \includegraphics[scale=0.5]{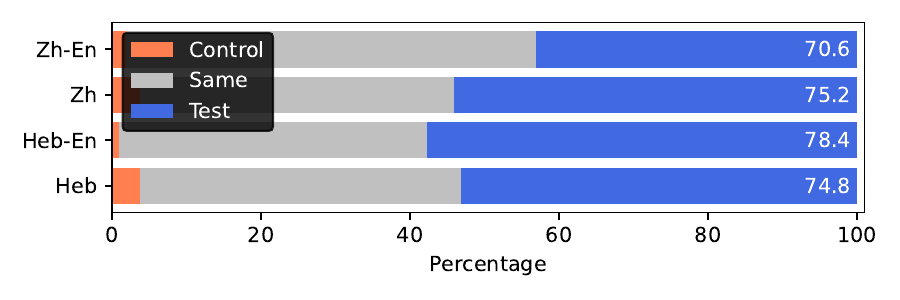}

    \caption{Human detection of semantic leakage in multilingual and crosslingual settings.}
    \label{fig:human_eval_multi}
\end{figure}

\autoref{fig:human_eval_multi} shows the human evaluation results for the multilingual and crosslingual experiments (with \lm{Gpt4o}). We can see significant semantic leakage in multilingual and crosslingual settings, with Leak-Rate values that range from 70.6 to 78.4 for the 4 different settings, similar to the Leak-Rate values we got for English.\footnote{In 6 examples of the crosslingual prompts (Zh-En), the generations are just translations because of the way the prompt is designed - we annotate those cases as ``neutral''.}

The generation quality with the \lm{Llama} model is generally much lower, and we get Leak-Rates of 66.5 and 61.5 for Chinese and Chinese-English, respectively, according to human annotation. As noted above, the quality of generation in Hebrew and Hebrew-English did not allow for evaluation of semantic leakage.

\autoref{tab:multi} shows the automatic evaluation of the multilingual and crosslingual experiments. The results for the multilingual settings are not as reliable as in the English setting since we cannot determine the quality of the underlying embedding methods for Hebrew and Chinese.\footnote{To the best of our knowledge, the model underlying sentenceBERT was trained (predominantly) on English. This is why the metric detects more leakage in cross-lingual settings (with evaluations on English) compared to other languages.} In the crosslingual setting we mainly compare English generations with English concepts, thus the results are more reliable.

\begin{table}[ht]
\centering
\resizebox{0.8\columnwidth}{!}{

\begin{tabular}{l|l|r|r|r}

& & \multicolumn{3}{c}{\textbf{Leak-Rate}} \\
\textbf{Model} & \textbf{Language} & BS & SB & OAI\\ \hline \hline

\multirow{4}{*}{GPT4o} & Heb & 60.6 & 53.5 & 67.2\\
& Heb-En & 62.1 & 58.2 & 74.9\\
& Zh & 67.6 & 48.9 & 80.8\\
& Zh-En & 61.6 & 60.5 & 71.3\\

\midrule

\multirow{2}{*}{3-70b-ins} & Zh & 73.4 & 54.1 & 82.8\\
 & Zh-En & 79.4 & 81.2 & 85.6\\

\bottomrule
\end{tabular}
}
\caption{Semantic leakage scores for multilingual and crosslingual setting, averaged across 10 samplings and measured by Leak-Rate.}
\label{tab:multi}
\end{table}

\section{Open-Ended Generation}
\label{sec:open}

We also examine semantic leakage in open-ended scenarios that encourage the model to generate multiple sentences or paragraphs. It is less clear how to quantify the semantic leakage in this setting, as it can manifest in more ways within the longer output; we therefore rely on more qualitative analysis in addition to the automatic metrics. This section focuses on generations from \lm{Gpt4o} with $t=1$, as this model with this temperature was most prone to leakage in previous experiments (Section \ref{sec:auto-results}). 

\paragraph{Stories} A popular setting for open-ended text generation is storytelling: we ask the model to write a story about a child, with the child's name being the concept we expect to leak. \autoref{fig:examples_stories} gives the story that \lm{Gpt4o} generates when prompted to create a story about a child named Coral (and also Melody in \autoref{fig:app_examples_stories} in the Appendix). These generated stories contain many words and elements directly related to the semantics of each name, and the narrative of each story is tailored to those meanings. 
When measuring semantic leakage in this setting automatically for 23 different names,\footnote{We remove any instance of the name from the generation, even when that instance is used as a common noun.} we get Leak-Rates of 78.7, 60.9 and 76.1 with BS, SB and OAI embeddings respectively, suggesting similar rates of leakage to the other examples.

\begin{figure}[ht!]
    \centering
    \resizebox{0.9\linewidth}{!}{
    \tcbset{colback=blue!8!white, colframe=bblue!75!black, 
    fonttitle=\bfseries, title=Semantic leakage in story generation, width=\linewidth+0.1cm, halign title=center }
    
    \begin{tcolorbox}
    Prompt: \textcolor{pcolor} {\textit{Tell me a short story about a child named \underline{Coral}.}} \\
    Model: \textcolor{gcolor} {\textit{Sure, here is a short story about a child named Coral: In the bustling \underline{coastal} town of \underline{Seaview}, ... and the \underline{waves} danced upon the \underline{shore}, ... Coral spent most of her days exploring the wonders of the \underline{seaside}. ... She grabbed her \underline{seashell}-collecting basket and marched down to the \underline{beach}, ... Along the way, she encountered friendly \underline{crabs}, playful \underline{dolphins} ... telling stories of the \underline{ocean's} many secrets. ... From that day on, Coral became known as the child who brought the \underline{ocean's} magic to the town. She shared the \underline{shells}, ... the mysteries of the \underline{sea} ...  }}

    \end{tcolorbox}  
    }
    \caption{Example of story generation, exhibiting semantic leakage based on the name in the prompt.}  
    \label{fig:examples_stories}
\end{figure}

\paragraph{Recipes} We also consider open-ended recipe generation. In~\autoref{fig:examples_recipes}, we present \lm{Gpt4o}'s generation when asked to suggest a recipe for use with a \textit{blue} pan (or a \textit{white} pan in \autoref{fig:app_examples_recipes} in the Appendix), where we consider each color to be the potentially leaking concept. The model suggests to make blueberry pancakes and chicken alfredo, respectively, with the foods matching the color of the pan. When prompted with a red pan, ingredients like peppers, tomatoes and paprika are far more frequent, same as with asparagus in the case of a green pan and lemon with a yellow pan. 

We experiment with 5 different colors (10 samples each) that are likely to be related to types of food (blue, green, red, white, and yellow). Automatic evaluation in this setting is more challenging since the concept might appear in the generation in a leaky way (``red pepper'') or just as part of the answer, with no attributed leakage (``red pan'') and these two cases are difficult to differentiate automatically. We decide to remove all occurrences of \prompt{<color> pan} from the generations when using automatic metrics, and get Leak-Rates of 74.0/58.0/74.0 with BS/SB/OAI embeddings, again suggesting semantic leakage here as well.

\begin{figure}[ht!]
    \centering
    \resizebox{0.9\linewidth}{!}{
    \tcbset{colback=blue!8!white, colframe=bblue!75!black, 
    fonttitle=\bfseries, title=Semantic leakage in recipe generation, width=\linewidth+0.1cm, halign title=center }
    
    \begin{tcolorbox}
    Prompt: \textcolor{pcolor} {\textit{I want to use my \underline{blue} pan, give me a recipe:}} \\
    Model: \textcolor{gcolor} {\textit{Sure! Here’s a recipe for a delicious \underline{Blueberry Pancake} that you can prepare using your blue pan: <recipe>}} 

    \end{tcolorbox}    
    }
    \caption{Example of recipe generation, exhibiting semantic leakage based on the color in the prompt.}  
    \label{fig:examples_recipes}
\end{figure}

\section{Related Work}

\citet{dalle-leakage} show similar conceptual leakage in image generation with \lm{Dalle-2} \cite{dalle2}, a text-to-image model by OpenAI. They find that sense-ambiguous words are hard for the model to isolate---similar to our observation that idioms in the prompt can leak the literal meaning into the generation. \citet{binding} try to mitigate this issue by encouraging the cross-attention maps to agree with the linguistic binding reflected by the prompt syntax. Leakage of properties in \lm{DALLE-2} has also been discussed by \citet{dalle2} and \citet{dalle2_syntax}.

As models have improved, a related line of research has asked whether they match human behavior, particularly with respect to cognitive biases and priming. Prior work compares models' behavior to documented behavior of humans in psychological experiments, highlighting some similarities as well as differences from human response to given information \cite{cog-bias-codex,cog-bias-llms,reasoning-bias-nature}. A known psychological phenomenon that is very related to models' semantic leakage as we present it here, is that of priming, and specifically semantic priming \cite{semantic-priming, semantic-priming2,priming-memory,priming-memory2}, where decisions and reactions about specific words change according to given semantic stimuli provided before the target word. We plan to investigate this relationship further in future work. 

\paragraph{Biases in Language Models}

There is an extensive line of work investigating bias and stereotypes of different demographic groups in NLP models, ranging from word embeddings \cite{bolukbasi,caliskan,lipstick} all the way to LMs \cite{stereoset,gender-kotek,persona-stereo}, sometimes focusing on specific end tasks \cite{mt-bias,mt-bias-gabi,bias-sentiment,coref-bias}. Existing works in this space usually investigate specific bias types with respect to concrete socio-demographic groups \cite{intersectional-bias,nationality-bias,disability-bias,bias-survey}, as well as toxicity in model generation \cite{toxicityprompts,honest}. Some works also point to specific input words that cause harmful generation, such as first names \cite{bias-bios,names} among others \cite{xstest,bias-list}. In this context, semantic leakage serves as a broader class of association bias, potentially pointing to the underlying mechanism that causes biases in model generation by latching onto concrete properties in the input, resulting in toxic or stereotyped generation.

\section{Conclusion}

We identified a phenomenon in generative language models never discussed before, which we term \textit{semantic leakage}: language models are prone to generating text that reflects unrelated semantic information from the prompt, leading to peculiar and sometimes unreasonable outputs.  We defined and measured semantic leakage in a range of models, and together with human evaluation show that it is prevalent and consistent across all models we test. We also found that semantic leakage occurs in many different generation settings, including multilingual and crosslingual ones. 

This leakage reflects associations learned by the model, similar to how different types of biases are learned; therefore, our characterization of semantic leakage broadens the scope we should consider with respect to potential ramifications of learned associations. While the implications of semantic leakage are not fully clear, it is a consistent behavior pattern in LMs that merits further study.

The finding that instruction-tuned models leak more is of special interest, given that they are the most popular and best performing model variations currently used. We hypothesize that semantic leakage is more dominant in these models because the leaking generations are less generic and seem to provide more information/content, which might be a property that is incentivized under these fine-tuning processes. We plan to explore this hypothesis more formally in future work.

\section*{Limitations}

While our experimental setup spans 13 models of different types and sizes, and explores different sampling temperature values, the scale of the prompts in our test suite remains limited due to the difficulty of manually creating prompts that are likely to leak in a way that we can detect and evaluate. This is also often the case when measuring other language model biases. 
In addition, though the results are consistent across all models and languages we experiment with, the trends might be different with other models or languages we have not tested. 

Finally, we cannot guarantee that the automatic evaluations do not include noise: in some cases, even after our automatic post-processing of the generations, our automatic metrics might consider non-leaking instance as leaking (e.g., in cases of occurrences of the concept in the generation due to a partial repetition of the prompt). However, the human evaluation we conduct generally agrees with the findings we get with the automatic detection, supporting their reliability.

\section*{Ethical Considerations}

Semantic leakage is a broad umbrella that may encompass many types of biases, and it broadens our perception and understanding about association biases that are being learned by models and then exhibited in generation. As such, semantic leakage might manifest at times as a concrete type of bias, e.g., cultural, gender or racial biases. This depends on the specific setting and the chosen concept that leaks; while our prompts and concepts are fairly general, these biases and stereotypes could potentially surface in our evaluation.\footnote{For example, some concepts in the evaluation involve food and music associated with specific countries or ethnicities.} 

However, our main goal in this paper is not to focus on specific types of biases (or biases that are particularly harmful) but to characterize and quantify the broader class of semantic leakage. We believe that a better understanding of the greater picture will in turn arm us with better tools to combat the subset of biases and behaviors that do have harmful consequences in practice.

Apart from that, in this paper the annotations and translations were done by several volunteers that were willing to help for the project. The load of this work was very low.

\section*{Acknowledgements}

This research received support through Schmidt Sciences, LLC. We thank Shauli Ravfogel and Ido Levin for helpful discussions and ideas. We also thank the translators and annotators for their contribution to the paper (Weijia Shi, Jacqueline He, Jacob Schreiber) and the ARK lab members for their valuable feedback. Finally, we would like to express our gratitude to the reviewers of this paper for their useful comments and suggestions.

\bibliography{anthology,custom}
\bibliographystyle{acl_natbib}

\newpage

\appendix

\section{Semantic Leakage in GPT Models}

\begin{figure}[ht!]
\centering

    \includegraphics[scale=0.5]{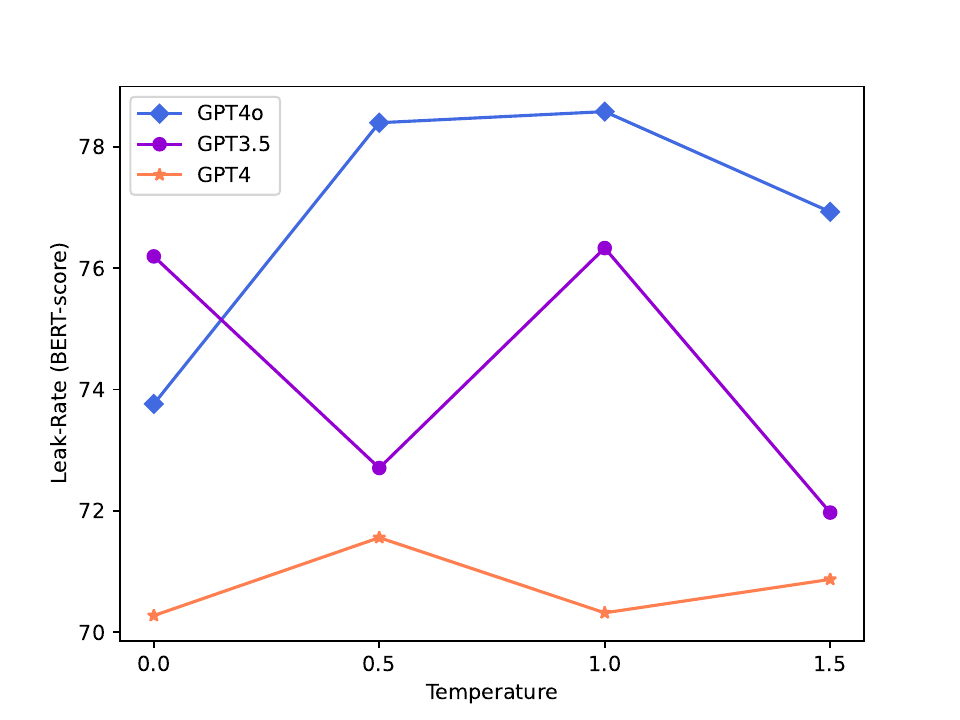}

    \caption{Semantic leakage in \lm{GPT} models using different temperature values (measured with Leak-Rate using BERT-score).}
    \label{fig:app_gpt4o}
\end{figure}

\autoref{fig:app_gpt4o} shows the leakage estimation in the three different \lm{GPT} models across different temperature values. \lm{Gpt4o} consistently leaks more than \lm{Gpt4} and \lm{Gpt-3.5}.

\label{sec:gpt4o_app}

\section{Multilingual and Crosslingual Semantic Leakage}

\autoref{tab:app_examples_multi} shows example multilingual and crosslingual prompts and generations.

\begin{table*}[ht]
\centering

    \includegraphics[scale=0.8]{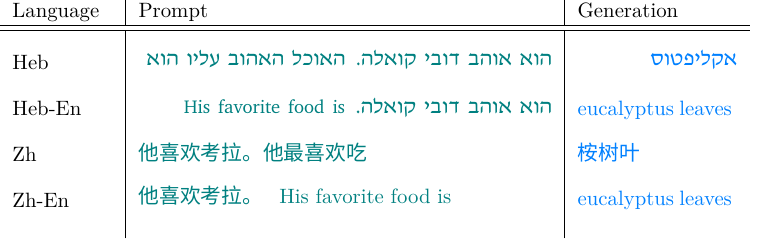}

    \caption{Example prompts and generations in Hebrew, Hebrew-English, Chinese and Chinese-English, by this order. All are translations of the prompt: \prompt{He likes koalas. His favorite food is}, and the Hebrew/Chinese generations can be translated to ``eucalyptus''/``eucalyptus leaves'', respectively.}
    \label{tab:app_examples_multi}
\end{table*}

\newpage

\section{Open-Ended Generation}

\autoref{fig:app_examples_stories} gives the story that \lm{Gpt4o} generates when prompted to create a story about a child named Melody. The generated story contains many words and elements directly related to the semantics of the name, and the narrative of the story is tailored to that meaning.

In~\autoref{fig:app_examples_recipes}, we present \lm{Gpt4o}'s generation when asked to suggest a recipe for use with a \textit{white} pan, where we consider the color to be the potentially leaking concept. The model suggests to make chicken alfredo, matching the color of the pan.

\begin{figure}[ht!]
    \centering
    \resizebox{0.9\linewidth}{!}{
    \tcbset{colback=blue!8!white, colframe=bblue!75!black, 
    fonttitle=\bfseries, title=Semantic leakage in story generation, width=\linewidth+0.1cm, halign title=center }
    
    \begin{tcolorbox}

    Prompt: \textcolor{pcolor} {\textit{Tell me a short story about a child named \underline{Melody}.}} \\
    Model: \textcolor{gcolor} {\textit{Once upon a time in the bustling town of \underline{Harmonyville}, a young girl named Melody lived with her family in a cozy, \underline{music-filled} home. Melody ... talent for \underline{playing the violin}. ... every time she \underline{drew her bow across the strings}, filling her home with enchanting \underline{melodies}. ... often brought her \underline{violin} with her. ... sorrowful \underline{tune} ... Melody followed the \underline{sound} ... bird \underline{sang} its heart out. ... decided to \underline{play her violin} for the bird. ... the \underline{instrument}, crafting a \underline{tune} ... The bird \underline{listened} intently, its spirits lifting with every \underline{note}. ... as the bird \underline{sang}, ... her enchanting \underline{performances} ... awaited her \underline{melody} ... always remembering that the true power of \underline{music} ... }} \\ 

    \end{tcolorbox}   
    }
    \caption{Example of story generation, exhibiting semantic leakage based on the name in the prompt.}  
    \label{fig:app_examples_stories}
\end{figure}

\begin{figure}[ht!]
    \centering
    \resizebox{0.9\linewidth}{!}{
    \tcbset{colback=blue!8!white, colframe=bblue!75!black, 
    fonttitle=\bfseries, title=Semantic leakage in recipe generation, width=\linewidth+0.1cm, halign title=center }
    
    \begin{tcolorbox}

    Prompt: \textcolor{pcolor} {\textit{I want to use my \underline{white} pan, give me a recipe:}} \\
    Model: \textcolor{gcolor} {\textit{Sure, I'd be happy to help you with a recipe that makes use of your white pan! Let's try a classic \underline{Chicken Alfredo}. White pans are often non-stick and ideal for making creamy dishes like this. Here’s a simple recipe: <recipe>}} \\

    \end{tcolorbox}    
    }
    \caption{Example of recipe generation, exhibiting semantic leakage based on the color in the prompt.}  
    \label{fig:app_examples_recipes}
\end{figure}

\end{document}